\documentclass[letterpaper]{article} 
\usepackage{aaai24}  
\usepackage{times}  
\usepackage{helvet}  
\usepackage{courier}  
\usepackage[hyphens]{url}  
\usepackage{graphicx} 
\usepackage{natbib}  
\usepackage{caption} 
\usepackage{amsmath}
\usepackage{amssymb}
\usepackage{booktabs}
\usepackage{multirow}
\usepackage{subcaption}
\usepackage{algorithmicx}
\usepackage[noend]{algpseudocode}
\usepackage[linesnumbered,ruled,vlined]{algorithm2e}

\newcommand{\papername}{BlindTuner}

\title{I can't see it but I can Fine-tune it: On Encrypted Fine-tuning of Transformers using Fully Homomorphic Encryption \footnote{Accepted for the presentation at PPAI@The 38th Annual AAAI Conference on Artificial Intelligence 2024}}
\author {
    Prajwal Panzade \textsuperscript{\rm 1},
    Daniel Takabi \textsuperscript{\rm 2},
    Zhipeng Cai \textsuperscript{\rm 1}
}
\affiliations {
    \textsuperscript{\rm 1}Georgia State University \\
    \textsuperscript{\rm 2}Old Dominion University \\
    ppanzade1@student.gsu.edu, takabi@odu.edu, zcai@gsu.edu
}

\usepackage{bibentry}

\begin{document}

\maketitle

\begin{abstract}
In today's machine learning landscape, fine-tuning pre-trained transformer models has emerged as an essential technique, particularly in scenarios where access to task-aligned training data is limited. However, challenges surface when data sharing encounters obstacles due to stringent privacy regulations or user apprehension regarding personal information disclosure. Earlier works based on secure multiparty computation (SMC) and fully homomorphic encryption (FHE) for privacy-preserving machine learning (PPML) focused more on privacy-preserving inference than privacy-preserving training. In response, we introduce BlindTuner, a privacy-preserving fine-tuning system that enables transformer training exclusively on homomorphically encrypted data for image classification. Our extensive experimentation validates BlindTuner's effectiveness by demonstrating comparable accuracy to non-encrypted models. Notably, our findings highlight a substantial speed enhancement of 1.5$\times$ to 600$\times$ over previous work in this domain.

\end{abstract}

\section{Introduction}
Transformers have become the state-of-the-art neural architecture in fields like natural language processing and computer vision, owing to their reliance on attention mechanisms and large-scale pre-training (Vaswani et al., 2017; Devlin et al., 2018; Touvron et al., 2021). With the growth of cloud computing, many machine learning as a service (MLaaS) platforms now provide pre-trained transformer fine-tuning capabilities. However, handling sensitive data on these platforms necessitates adherence to privacy laws like GDPR (General Data Protection Regulation) and CCPA (California Consumer Privacy Act) (Ribeiro et al., 2015). In a standard MLaaS system, the client own the data while ML computation occurs on the cloud (Liu et al., 2021). When data contains private information like healthcare or banking records, significant privacy risks arise. Consider a hospital using an MLaaS cancer prediction service: while patients may consent to data access, externally sharing sensitive information for modeling can enable breaches and unauthorized access, violating privacy. Moreover, if patients revoke their consent, model development becomes difficult.

Many approaches to privacy-preserving machine learning using secure multiparty computation exist, including SecureML (Mohassel et al., 2018), SecureNN (Wagh et al., 2019), and DeepSecure (Rouhani et al., 2018). However, these techniques require substantial client-server communication (Wagh et al., 2019). For lower communication, fully homomorphic encryption techniques like CryptoNets (Gilad-Bachrach et al., 2016), CryptoDL (Hesamifard et al., 2018), and ML Confidential (ML Confidential, 2021) are preferred, though more research has focused on private inference over training (Reagen et al., 2021; Gilad-Bachrach et al., 2016). Recent work has proposed encrypted neural network training using FHE (Nandakumar et al., 2019). Moreover, privacy-preserving transfer learning methodologies are developed in Glyph (Lou et al., 2020) and HETAL (Lee et al., 2023). However, the existing methodologies used for privacy-preserving transfer learning can be further refined, specifically aiming to improve accuracy and reduce training duration. This paper focuses on addressing these refinements.

The contributions of this paper are as follows: 
\begin{itemize}
    \item We introduce BlindTuner, an optimized system designed for training transformer models using FHE-encrypted data, facilitating privacy-preserving fine-tuning within cloud environments.
    \item In an effort to develop privacy-preserving strategies for model training, our investigation delves into the complex relationship between FHE and data-efficient image transformers (DEiT).
    \item By conducting thorough experimentation and rigorous evaluation on a wide range of datasets (including MNIST, CIFAR-10, DermaMNIST, and face mask detection), we propose a methodology that guarantees accessibility without precluding the need for specialized knowledge in cryptography. We'll provide an open-source implementation of {\papername} at \url{https://github.com/prajwalpanzade}.
    \item Significantly, our implementation exhibits superior performance compared to Glyph (NeurIPS 2020), progressing by 300 times in the classification of MNIST digits and 600 times in the classification of DermaMNIST images.

\end{itemize}

\section{Background Knowledge}
\subsection{Fully Homomorphic Encryption}
Fully Homomorphic Encryption (FHE) stands as an advanced cryptographic technique, pioneered by Gentry \cite{gentry2009fully}, enabling computations to be executed on encrypted data without necessitating decryption. In essence, it permits a third party to perform computations on encrypted data without exposure to the data itself or the resultant calculations. The implications of FHE are profound for privacy and security across diverse applications, notably in cloud computing and data outsourcing scenarios.

FHE encompasses several variations, including CKKS (Cheon-Kim-Kim-Song) \cite{cheon2017homomorphic}, Brakerski-Gentry-Vaikuntanathan (BGV) \cite{brakerski2014leveled}, and the Brakerski/Fan-Vercauteren scheme (BFV). Among these, CKKS gains popularity owing to its adeptness in handling real numbers. Analogous to public key encryption (PKE), the CKKS scheme encompasses encryption, decryption, and key generation algorithms. However, distinct from PKE, CKKS integrates homomorphic addition and multiplication functionalities, enabling addition or multiplication of two ciphertexts.

A succinct overview of these algorithms is as follows:

\begin{itemize}
\item KeyGen(1$^\lambda$): Generates a public key (pk), secret key (sk), and an evaluation key (evk).
\item Enc$_{pk}$(m): Encrypts a message (m $\in$ R) using pk, producing ciphertext c, where R represents a set of real numbers.
\item Dec$_{sk}$(c): Outputs m for a given ciphertext c using sk.
\item Add(c$_1$,c$_2$): Outputs element-wise addition Enc(m$_1$+m$_2$) when provided with c$_1$ and c$2$.
\item Mult$_{evk}$(c$_1$,c$_2$): For a ciphertext pair (c$_1$,c$_2$) and an evk, generates element-wise multiplication Enc(m$_1$*m$_2$). Notably, both addition and multiplication yield ciphertexts, necessitating sk for decryption. ML computations rely on multivariate polynomials, and the CKKS scheme supports bootstrapping, facilitating the computation of multivariate polynomials of arbitrary degrees \cite{cheon2018bootstrapping}. The state-of-the-art encrypted matrix multiplication proposed by \cite{lee2023hetal} uses the basic FHE operations to optimize matrix multiplication algorithms of the form AB$^T$ and A$^T$B. Also, they proposed an optimized softmax approximation algorithm for ML computation. Both of these encrypted matrix multiplications and softmax approximations are the basis of our approach.
\end{itemize}
For additional details regarding the CKKS scheme, we suggest referring to \cite{cheon2017homomorphic} and \cite{cheon2018bootstrapping} for comprehensive insights and in-depth discussions on this encryption technique.
\subsection{Data-Efficient Image Transformers}
Vision transformers (ViT) \cite{dosovitskiy2020image} have emerged as a promising architecture for image classification tasks. However, historically, ViT models have necessitated extensive pre-training on large datasets to achieve competitive performance, unlike convolutional neural networks (CNNs) \cite{lecun1998gradient}. Touvron et al. introduced Data-efficient Image Transformers (DeiT) \cite{touvron2021training}, demonstrating that transformers can match or surpass state-of-the-art CNNs when trained exclusively on ImageNet. The authors employed various modifications to the training approach, incorporating extensive data augmentation techniques such as RandAugment, CutMix, and repeated augmentation. Additionally, they introduced a novel distillation procedure utilizing a distillation token that engages with other embeddings through self-attention. Without distillation, DeiT achieves a top-1 accuracy of 81.8\% on ImageNet, surpassing prior ViT models at 77.9\%. Employing distillation from a RegNet teacher, DeiT achieves an accuracy of 85.2\%, exceeding CNNs like EfficientNet \cite{touvron2021training}. Transfer learning to other datasets also reveals competitive performance, underscoring the generalization capability of vision transformers. Overall, DEiT signifies a substantial leap in establishing transformers as a credible alternative to CNNs in computer vision tasks. Notably, DEiT enables the training of high-performing transformer models without relying on extensive datasets.

\subsection{Transfer Learning}

Transfer Learning (TL) stands as a machine learning technique leveraging knowledge acquired from solving one problem to address a related but distinct problem \cite{pan2009survey}. Specifically, TL involves retraining a model initially trained on a large, general dataset using a second, smaller dataset \cite{weiss2016survey}. The rationale behind TL lies in the notion that lower levels of neural networks identify fundamental, universal features such as edges, which hold relevance across various tasks. Consequently, pre-trained models serve as valuable starting points, necessitating less data to learn task-specific features. In computer vision, TL extensively employs large models like VGG \cite{simonyan2014very}, ResNet \cite{he2016deep}, and EfficientNet \cite{tan2019efficientnet}, pre-trained on ImageNet \cite{deng2009imagenet}, and subsequently fine-tuned for specialized domains like medical or aerial imaging. By capitalizing on pre-trained features, TL achieves high accuracy using moderately sized datasets. The process of fine-tuning a pre-trained model proves notably faster and more data-efficient compared to training a model from scratch. Fine-tuning remains a predominant methodology responsible for breakthrough advancements, particularly in applications constrained by limited training data.

\section{The Proposed Methodhology}
\label{proposed}
\subsection{Threat Model and overview}
\begin{figure*}[!ht]
    \centering
    \includegraphics[scale=0.55]{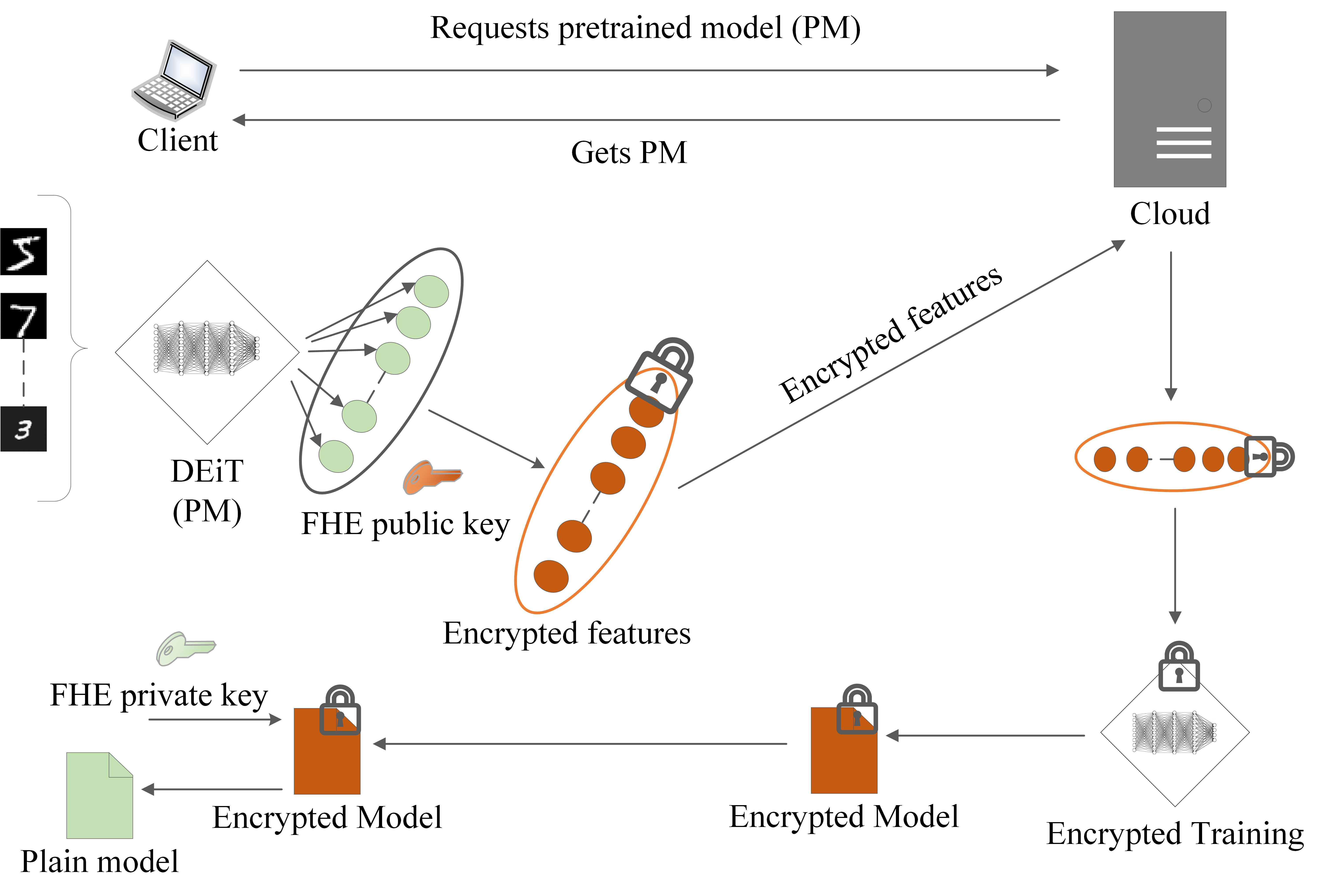}
    \caption{Overview of the proposed end-to-end \papername}
    \label{blindTL}
\end{figure*}

Our system, {\papername}, depicted in Figure \ref{blindTL}, involves two primary entities: the client and the cloud. The client, as the data owner, transmits FHE-encrypted data to the cloud, which is responsible for conducting machine learning computations in an MLaaS environment. The cloud executes the ML computations, generating an encrypted trained ML model, thus safeguarding the client's data privacy. Operating under an honest but curious security model \cite{goldreich2004foundations}, {\papername} ensures compliance with the protocol while both cloud and client actively seek maximum information. Leveraging the fine-tuning of pre-trained models, {\papername} assumes a shared pre-trained model for feature extraction between the client and the cloud. Subsequent subsections delve into the intricacies of computations carried out by both entities.
\subsection{{\papername}}
Algorithm 1 presents the comprehensive sequence of actions executed by both the client and the cloud within the {\papername} system.

\begin{algorithm}[!ht]
\caption{The {\papername} System}
\KwIn{Training data and labels $(X_{\text{train}}, Y_{\text{train}})$, Validation data and labels $(X_{\text{val}}, Y_{\text{val}})$}
\KwOut{Trained model $W_{\text{pt}}$}

\textbf{Step 1: Feature Extraction}\;
The client requests the cloud for the DEiT model tailored for image classification tasks\;
Using the DEiT model, the client extracts features from $(X_{\text{train}}, Y_{\text{train}})$ and $(X_{\text{val}}, Y_{\text{val}})$\;

\textbf{Step 2: Data Encryption}\;
The client encrypts $(X_{\text{train}}, Y_{\text{train}})$ as $D_{\text{ct\_train}}= \{(X_{\text{ct\_train}}, Y_{\text{ct\_train}})\}$ and $(X_{\text{val}}, Y_{\text{val}})$ as $(X_{\text{ct\_val}}, Y_{\text{ct\_val}})$ using the FHE public key and transmits to the cloud\;

\textbf{Step 3: Training Loop}\;
\For{each epoch}{
  \textbf{a) Parameters update with NAG}\;
  \For{each $(X_{\text{ct}}, Y_{\text{ct}})$ in $D_{\text{ct\_train}}$}{
    The cloud updates parameters using NAG:\;
    $W_{\text{ct}}^{t+1} = V_{\text{ct}}^t - (\alpha /N) (P_{\text{ct}} - Y_{\text{ct}})^T X_{\text{ct}}$\;
    $V_{\text{ct}}^{t+1} = (1 - \gamma_t)W_{\text{ct}}^{t+1} + \gamma_t W_{\text{ct}}^t$\;
  }
  where $\alpha$ = learning rate, $N$= batch size, $P_{\text{ct}} = \text{ASoftmax}(X_{\text{ct}}(V_{\text{ct}}^t)^{\intercal})$ with ASoftmax, and $\{\gamma_t\}_{t\geq0}$ as proposed in \cite{lee2023hetal}\;
  \textbf{b) Compute and Send Logits}\;
  The cloud computes logits $\ell_{\text{ct\_val}} = X_{\text{ct\_val}}(W_{\text{ct}})^T$ and transmits them to the client\;

  \textbf{c) Decryption and Validation Loss Computation}\;
  The client decrypts logits and computes the validation loss $L_{CE}$ with $\ell_{\text{pt\_val}}$ and $Y_{\text{pt\_val}}$\;
}

\textbf{Step 4: Finalization}\;
The cloud transmits $W_{\text{ct}}$ to the client, which decrypts it using the FHE private key to obtain the plain model $W_{\text{pt}}$\;

\end{algorithm}

\subsubsection{Client}
The client, as the data owner, seeks privacy-preserving model training services from the cloud to offload computational resources. Initially, both the client and the cloud agree upon using a DEiT pre-trained model for fine-tuning. The client utilizes DEiT to extract features from their dataset intended for model training. Post feature extraction, the client preprocesses and encrypts the features using CKKS-based FHE, employing its public key, resulting in encrypted features. These encrypted features are subsequently transmitted to the cloud for further processing.

\subsubsection{Cloud}
Upon receiving the encrypted features from the client, the cloud employs these encrypted features for fine-tuning the ML model. The fine-tuning process integrates Nesterov's accelerated gradient (NAG) \cite{nesterov1983method} and encrypted matrix multiplication, as proposed in \cite{lee2023hetal}, for softmax approximation. NAG is a renowned method facilitating faster convergence in homomorphic encryption-based ML computations \cite{crockett2020low}. Since all computations occur on encrypted data, no data is exposed to the cloud. Post fine-tuning, the trained ML model is prepared for inference. During inference, only encrypted features are necessary, and the cloud forwards the output layer results back to the client. The client subsequently decrypts the results using their private key.

\subsection{Security Analysis}
Here, we evaluate the security of {\papername}. During fine-tuning, the cloud exclusively handles encrypted client features, ensuring no data exposure, even in an adversarial context. Regarding potential attacks on FHE cryptosystems, these systems are regarded as quantum-secure, safeguarding original plaintext data and computed outcomes against unauthorized alterations, even in compromised infrastructure scenarios \cite{creeger2022rise}.

\section{Experimental Results}
\subsection{Datasets}
\label{datasets}
To showcase our experimental evaluations in image classification, we utilized four well-known datasets.

\subsubsection{MNIST}
The MNIST dataset \cite{lecun1998gradient} comprises 28x28 pixel grayscale images of handwritten digits ranging from 0 to 9, often used as a standard benchmark in computer vision research.

\subsubsection{CIFAR-10}
CIFAR-10 \cite{krizhevsky2009learning} consists of 32x32 color images categorized into ten classes, commonly employed for image classification tasks.

\subsubsection{DermaMNIST from MedMNIST}
The DermaMNIST dataset \cite{tschandl2018ham10000} is a subset of MedMNIST \cite{yang2023medmnist}, featuring grayscale images portraying various skin conditions segregated into seven classes.

\subsubsection{Face Mask Detection}
The face mask detection dataset \cite{makeml} includes images of individuals both with and without face masks and masks worn incorrectly, presenting a 3-class classification problem.

\subsection{Environment Setup}
Our primary computational tool for FHE operations is the HEaaN library \cite{cheon2017homomorphic}, which supports bootstrapping \cite{cheon2018bootstrapping}. We utilize the GPU library version of HEaaN obtained from the developers of HEaaN\footnote{\url{https://heaan.it/}}. For feature extraction using pre-trained transformers on the client side, we rely on PyTorch, TorchVision, NumPy, and transformers libraries. Our experiments are conducted on a workstation equipped with an Intel Xeon Gold 6230 R processor, operating at a clock speed of 2.10 GHz, and 755 GB of usable RAM, NVIDIA Tesla V100 32 GB GPU, running the Ubuntu operating system. Specific software versions used in our experiments are detailed in Table \ref{table:swversions}.
\begin{table}[!ht]
\caption{Software specifications}
\begin{center}
\begin{tabular}{cc}
\hline
\cline{1-2} 
 \textbf{Software} & \textbf{Version}\\
\midrule
Ubuntu & 20.04.4 \\
Python & 3.8.5\\
Transformers & 4.33.1\\
Numpy & 1.22.2\\
PyTorch & 2.0.1 \\
Torch Vision & 0.15.2 \\
HEaaN & 0.2.0 \\
\hline
\end{tabular}
\label{table:swversions}
\end{center}
\end{table}

\subsection{Experiments}
\subsubsection{\textbf{The Client}}
The client utilizes a DEiT pre-trained model to extract features from the specified datasets as detailed in the previous section. Four benchmark datasets—MNIST, CIFAR-10, Face Mask Detection, and DermaMNIST—are employed for rigorous evaluation. We leverage the deit-base-distilled-patch16-224 version of the DEiT model by Facebook research, accessible through the Transformers library. Post feature extraction, the client partitions the dataset into train, validation, and test sets. Before transmitting the training and validation sets to the cloud, they are encoded and encrypted using the ML submodule of the HEaaN library. The FHE context setting utilized is FGb, featuring a cyclomatic ring dimension of 2$^{16}$, ensuring a 128-bit security level, as described in \cite{cheon2021practical}. Keys are generated individually for each dataset experiment, generated at the experiment's outset and maintained throughout. Consistency in context settings is ensured across all experiments. Feature extraction time in the plain domain is not factored in, as it's considered an offline process and doesn't impact the encrypted fine-tuning demonstration.

\subsubsection{\textbf{The Cloud}}
Upon receiving the encrypted training and validation sets, the cloud sets up the ML model for encrypted training using the ML submodule of the HEaaN library. Hyperparameter tuning begins with a batch size of 2048, a learning rate of 1, and 10 epochs. After multiple hyperparameter modifications, optimal configurations are determined, detailed in Table \ref{tab:parameters}. Experimental results, provided in Table \ref{tab:results}, showcase the outcomes of experiments on various datasets. Enc training time denotes the duration for encrypted model training, while Enc accuracy signifies the test accuracy of the encrypted model. Similarly, Unenc accuracy and Unenc time represent test accuracy and computation time for the unencrypted model. Identical hyperparameters are used for both domains to facilitate comparison. Results (Table \ref{tab:results}) highlight the performance of {\papername} for both encrypted and unencrypted models, with slight disparities observed in encrypted model performance compared to their unencrypted counterparts. This demonstrates the effectiveness of approximation arithmetic proposed in \cite{cheon2017homomorphic} and \cite{lee2023hetal} for accurate ML model training, despite the cryptographic FHE computations' time-intensive nature. Nonetheless, the benefits of preserving user data privacy without revealing information to the cloud outweigh the computational time. Moreover, {\papername} exhibits competitive performance compared to prior privacy-preserving fine-tuning
\begin{figure*}[!ht] 
    \centering
    \begin{subfigure}{0.497\textwidth}
        \includegraphics[width=\linewidth]{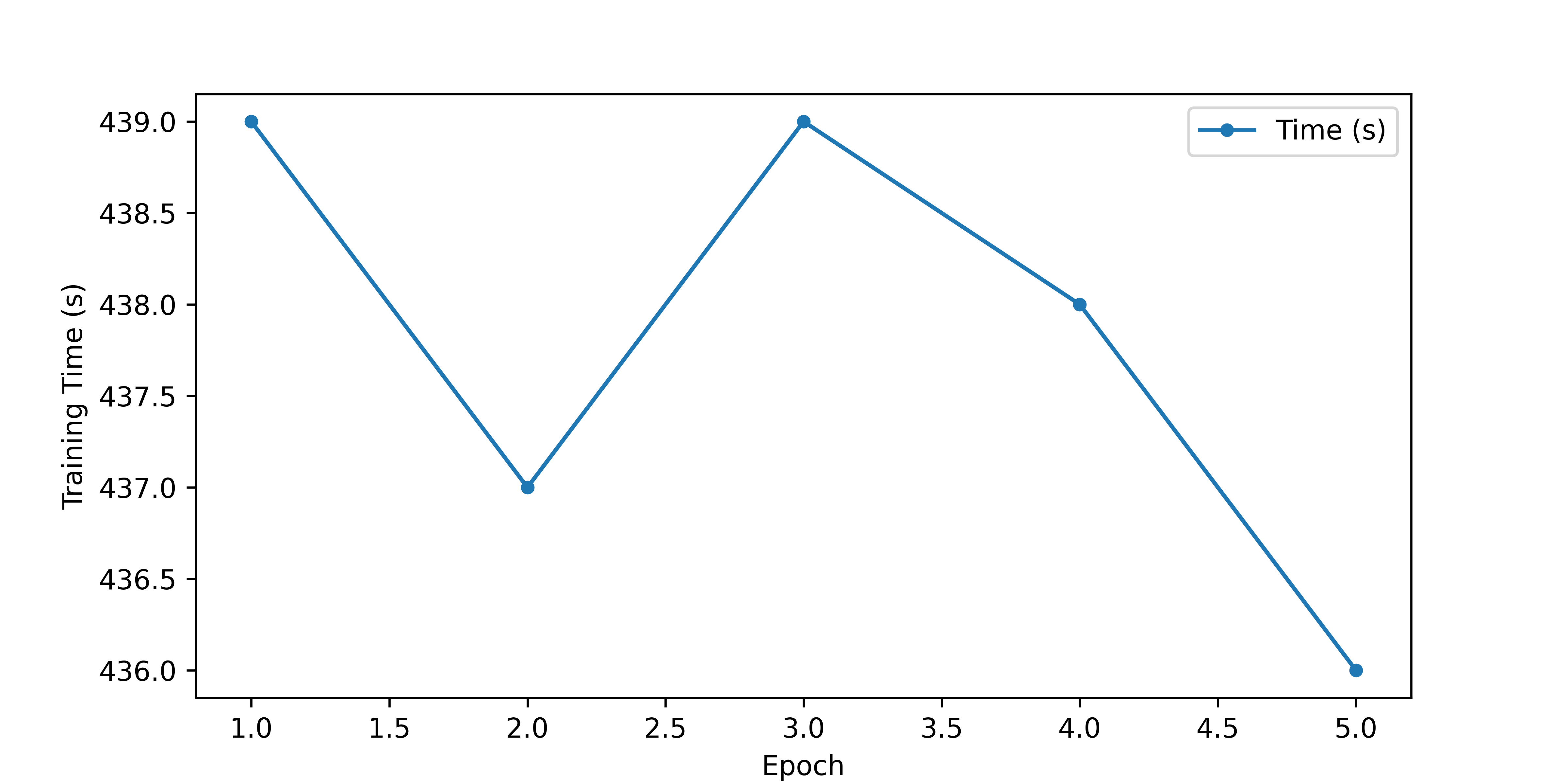}
        \caption{MNIST}
        \label{subfig:sub1}
    \end{subfigure}
    \hfill
    \begin{subfigure}{0.497\textwidth}
        \includegraphics[width=\linewidth]{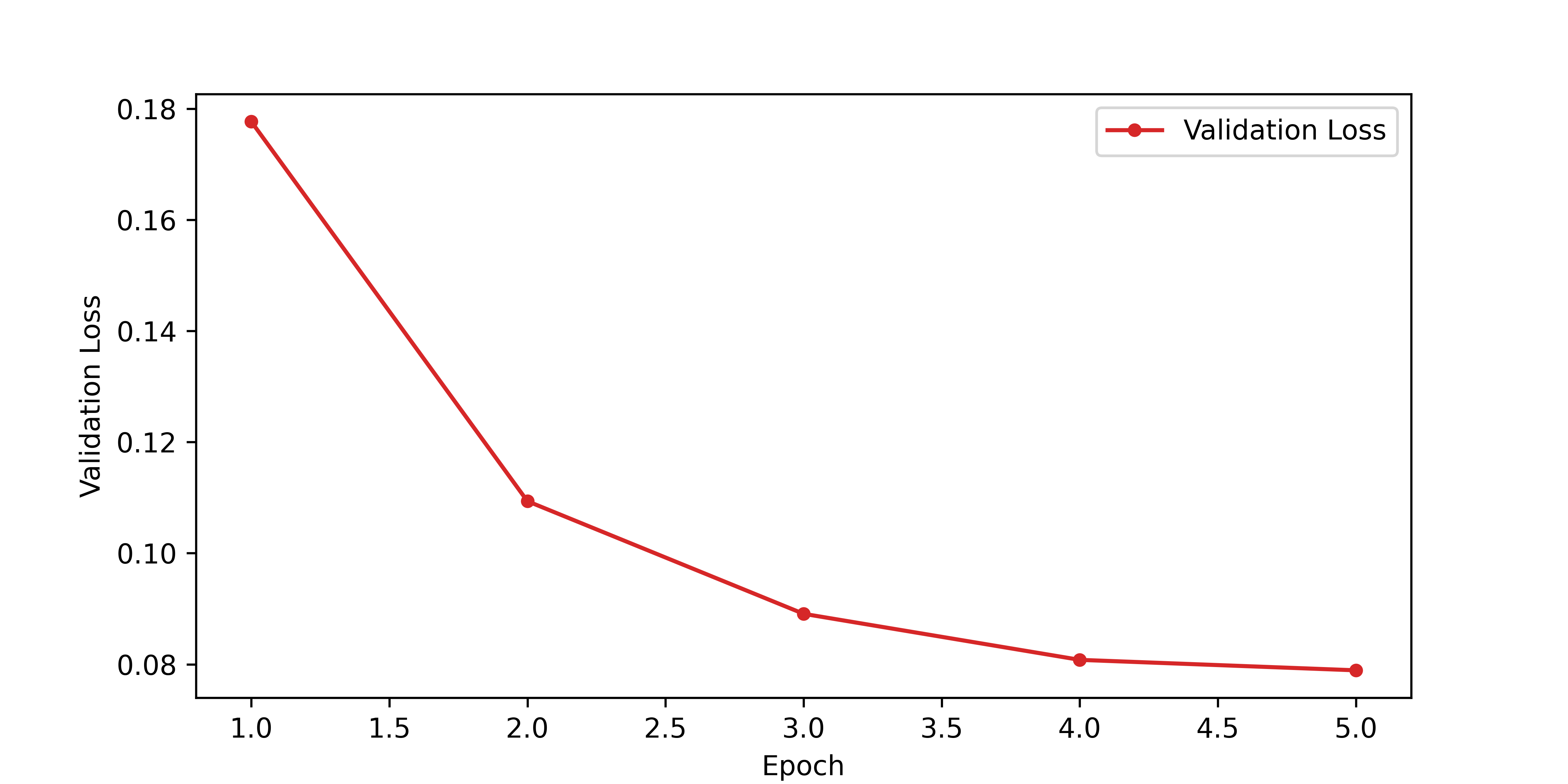}
        \caption{MNIST}
        \label{subfig:sub2}
    \end{subfigure}
    
    \begin{subfigure}{0.497\textwidth}
        \includegraphics[width=\linewidth]{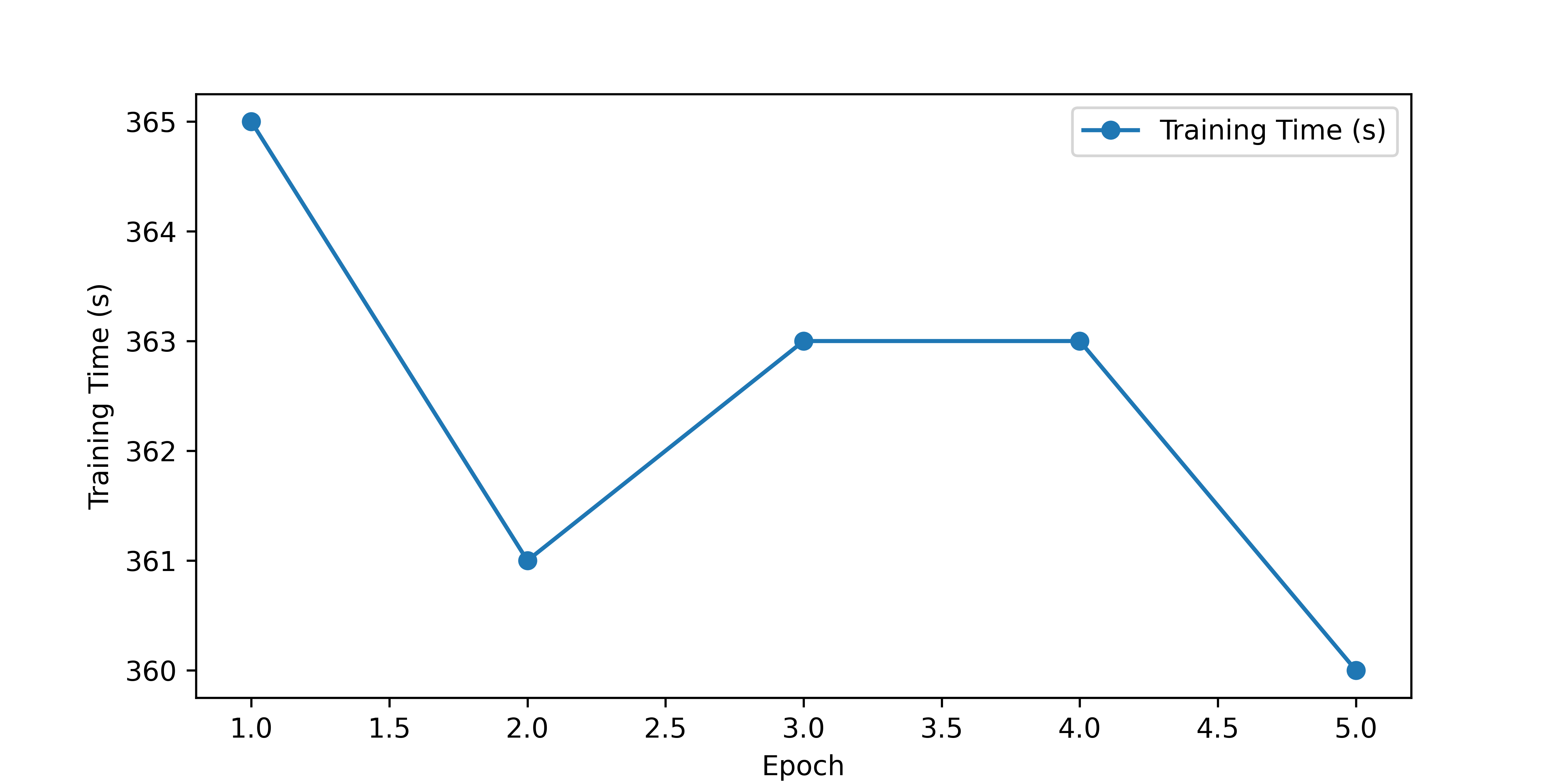}
        \caption{CIFAR-10}
        \label{subfig:sub1}
    \end{subfigure}
    \hfill
    \begin{subfigure}{0.497\textwidth}
        \includegraphics[width=\linewidth]{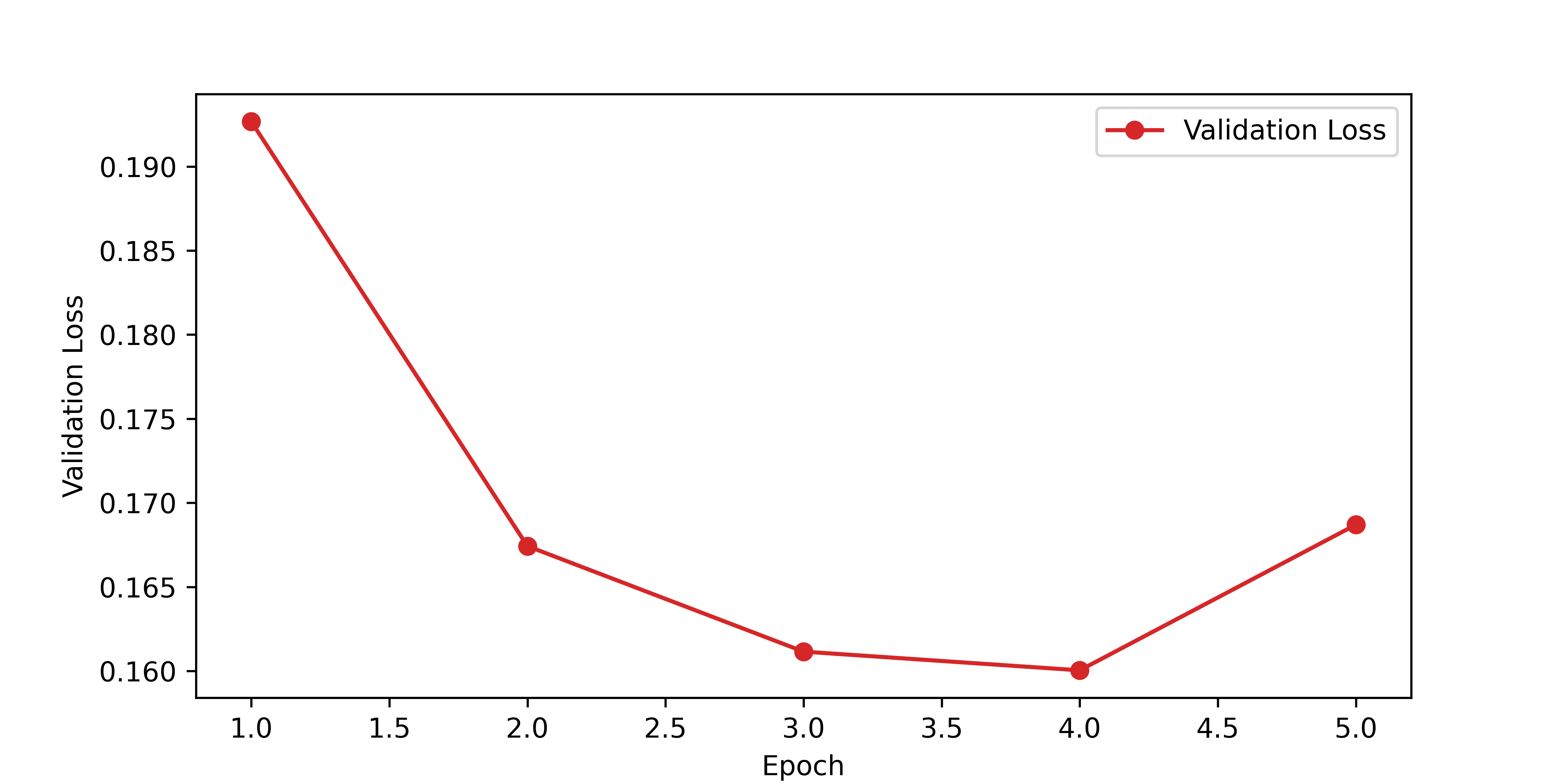}
        \caption{CIFAR-10}
        \label{subfig:sub2}
    \end{subfigure}
    \vspace{10pt} 
    \begin{subfigure}{0.497\textwidth}
        \includegraphics[width=\linewidth]{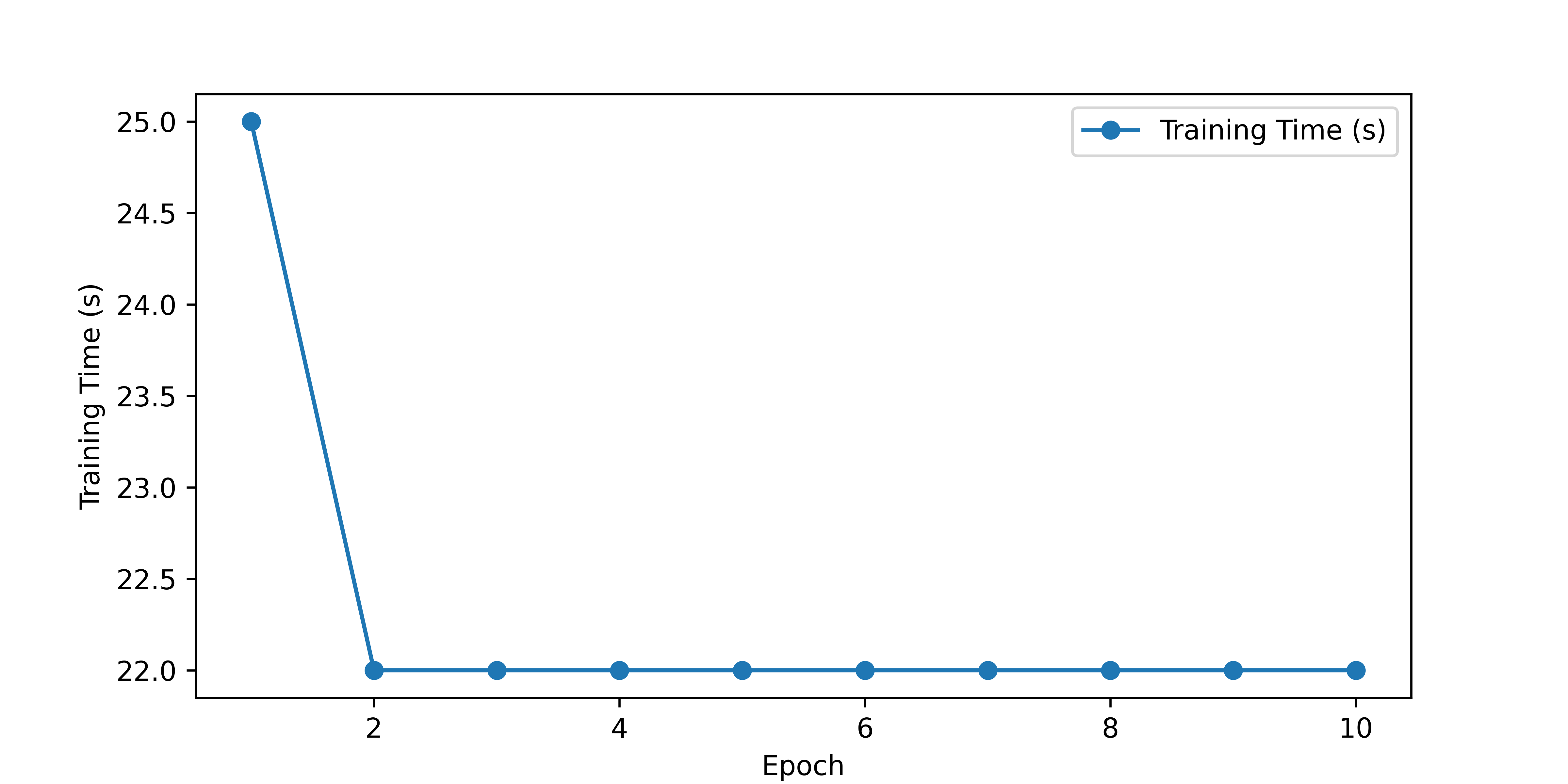}
        \caption{Face Mask Detection}
        \label{subfig:sub3}
    \end{subfigure}
    \hfill
    \begin{subfigure}{0.497\textwidth}
        \includegraphics[width=\linewidth]{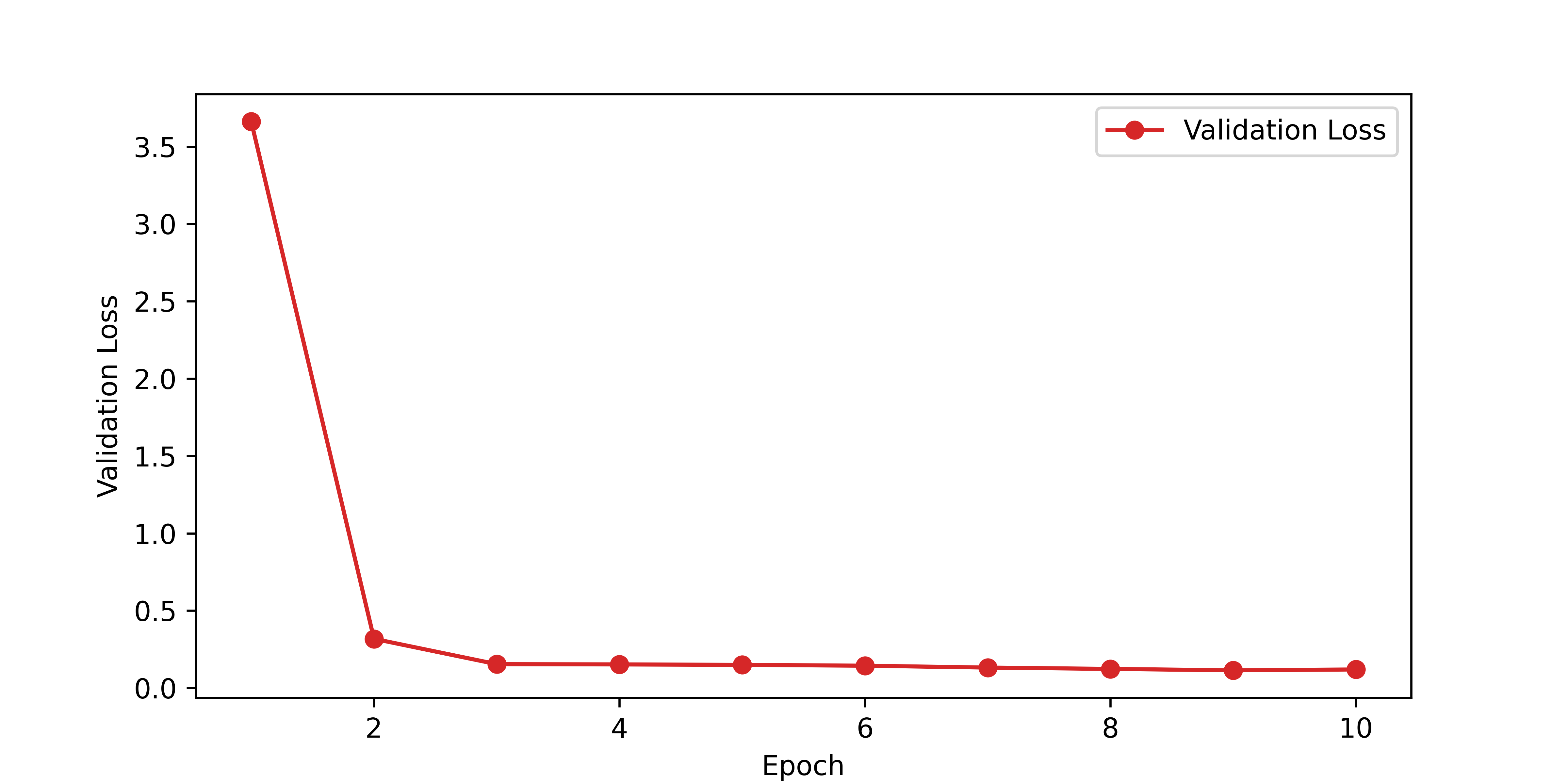}
        \caption{Face Mask Detection}
        \label{subfig:sub4}
    \end{subfigure}

    \begin{subfigure}{0.497\textwidth}
        \includegraphics[width=\linewidth]{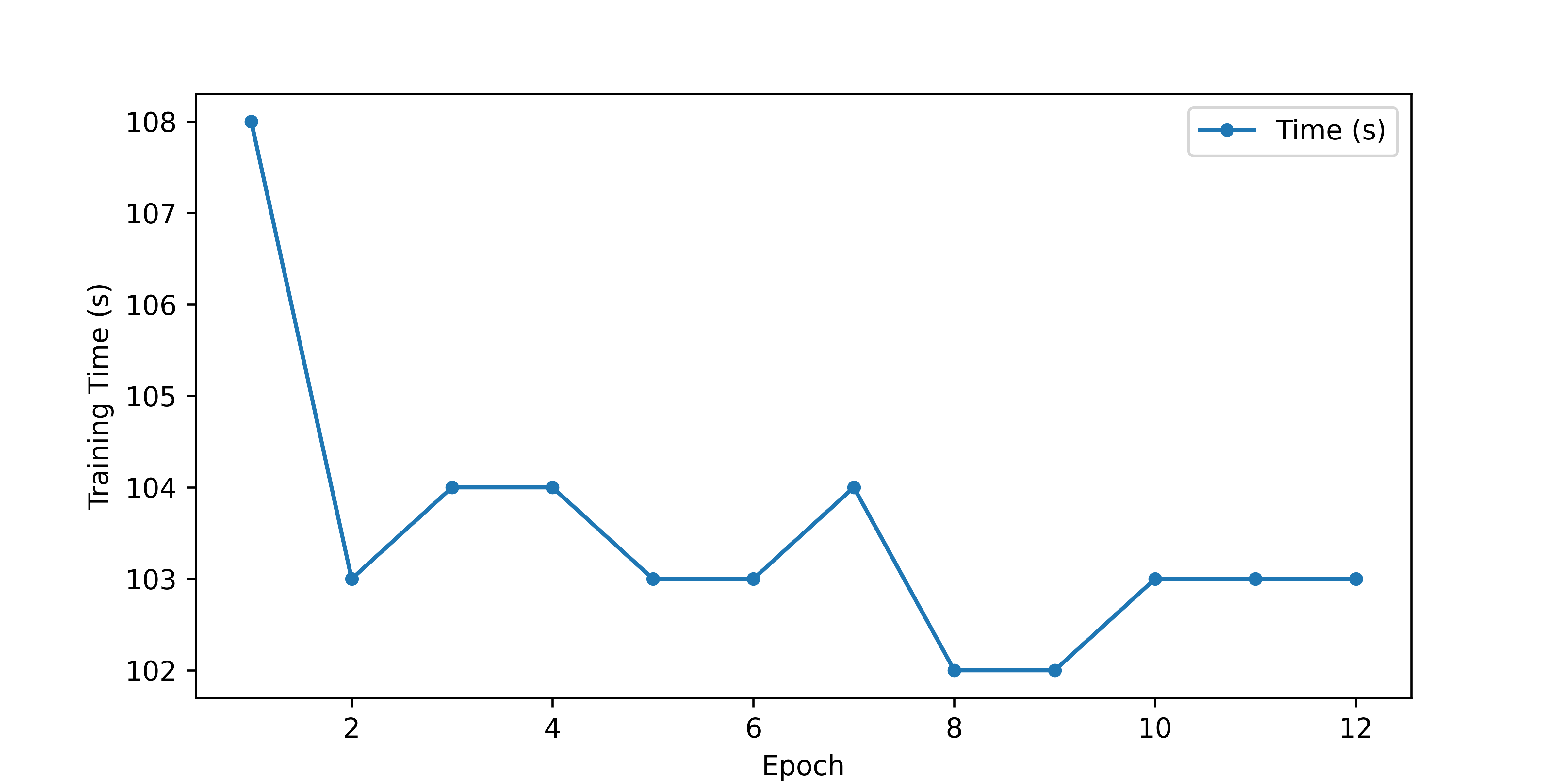}
        \caption{DermaMNIST}
        \label{subfig:sub3}
    \end{subfigure}
    \hfill
    \begin{subfigure}{0.497\textwidth}
        \includegraphics[width=\linewidth]{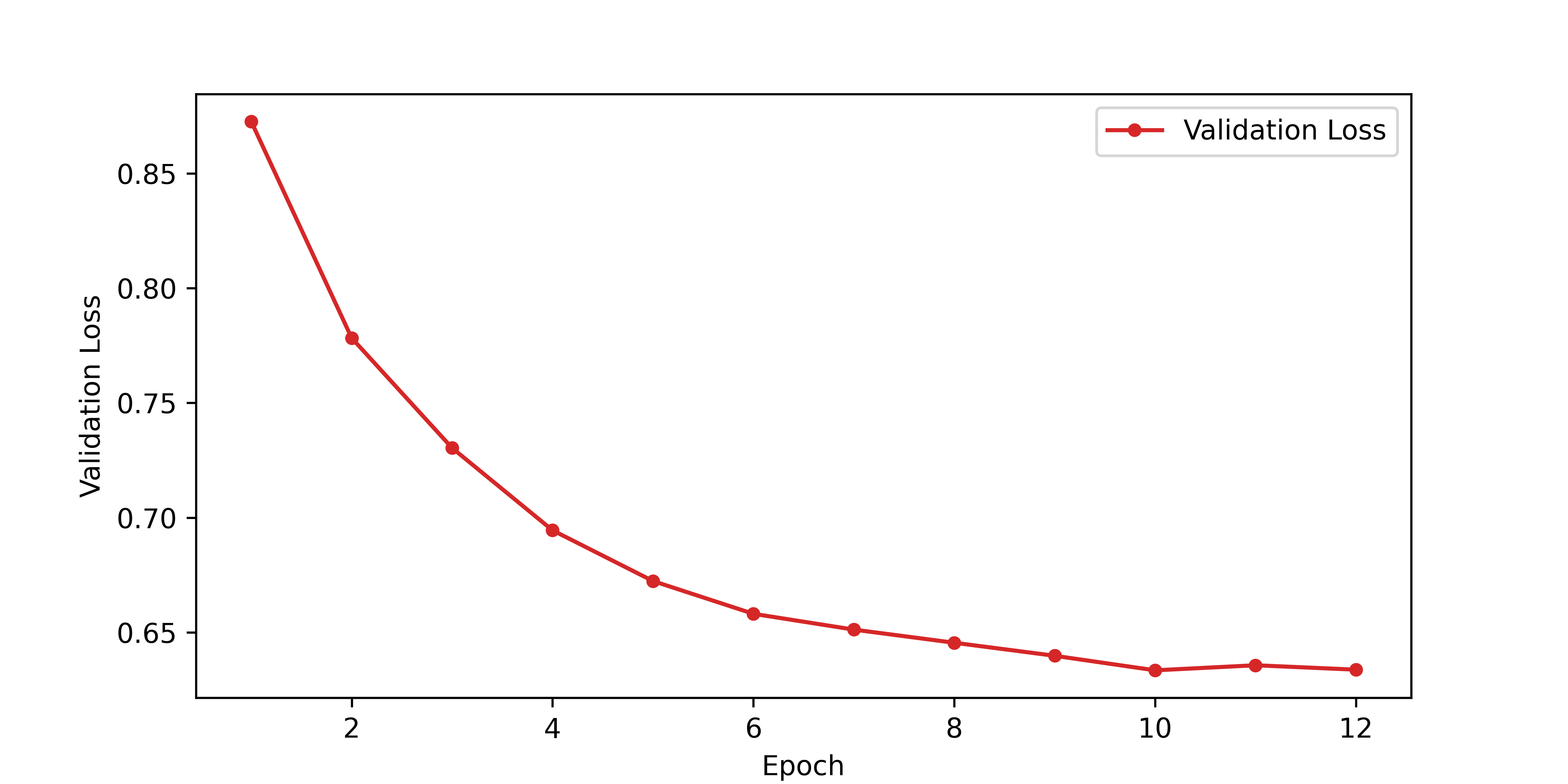}
        \caption{DermaMNIST}
        \label{subfig:sub4}
    \end{subfigure}
    \caption{{\papername} performance on various datasets}
    \label{fig:main}
\end{figure*}on GPUs. A comparative analysis with earlier works is discussed in the subsequent subsection.
\subsection{Comparison}
Table \ref{table:comparison} presents a comprehensive comparison of the training latency between our proposed approach, {\papername}, and
\begin{table*}[!ht]
\caption{Performance of \papername}
\begin{tabular}{cccccc}
\toprule
 \textbf{Dataset} & \textbf{Enc training time (min)} & \textbf{Enc Accuracy} & \textbf{Unenc accuracy} & \textbf{Unenc time (s)}  & \textbf{\#Epochs} \\
\midrule
    MNIST & 37.8390 & 97.75\% & 98.02\% & 44.74 & 5 \\
        CIFAR-10 & 31.3342 & 95.20\% & 95.20\% & 61.26 & 5 \\
        Face mask detection & 3.9086 & 97.18 \% & 97.06\% & 9.90 & 10 \\
        DermaMNIST & 16.2993 & 76.11\% & 76.16\% & 30.25 & 12 \\
\bottomrule
\end{tabular}
\label{tab:results}
\end{table*}
\begin{table*}[!ht]
\caption{Training parameters}
\centering
\begin{tabular}{cccc}
\toprule
\textbf{Dataset} & \textbf{\#Epochs} & \textbf{Learning rate} & \textbf{Batch size} \\
\midrule
MNIST & 5 & 0.1 & 1024 \\

CIFAR-10 & 5 & 0.1 & 1024 \\
Face mask detection & 10 & 0.1 & 512 \\
DermaMNIST & 12 & 0.01 & 512 \\
\hline
\end{tabular}
\label{tab:parameters}
\end{table*}two existing methodologies, Glyph \cite{lou2020glyph} and HETAL \cite{lee2023hetal}. In the interest of a fair assessment, we conduct an evaluation of HETAL on our workstation, while relying on the reported results for Glyph due to its similar architecture to ours.

Analyzing the MNIST dataset results, {\papername} achieves a commendable test accuracy of 97.75\%, nearly on par with Glyph's 98.6\%. However, {\papername} significantly reduces training time to a mere 37.839 minutes, a drastic improvement compared to Glyph's extensive 8-day requirement. This renders {\papername} over 300 times faster than Glyph. Compared to HETAL, {\papername} exhibits superior performance by enhancing accuracy from 96.73\% to 97.75\%, concurrently reducing training time from 57.3715 to 37.839 minutes, demonstrating a 1.5 times increase in speed while improving accuracy.

For the CIFAR-10 dataset, {\papername} achieves a competitive test accuracy of 95.20\% in 31.3342 minutes, remaining competitive with HETAL, which achieves a slightly higher accuracy of 96.57\% but necessitates a longer training duration of 51.905 minutes. Overall, {\papername} showcases robust performance in CIFAR-10 relative to HETAL, achieving comparable accuracy with faster training.

On the Face Mask Detection dataset, {\papername} achieves an impressive accuracy of 97.18\% in just 3.9086 minutes, surpassing HETAL by 2.4 times in training efficiency.

Examining the DermaMNIST dataset, Glyph reports an accuracy of 73.2\% with a substantial training period of 7 days. In contrast, {\papername} maintains competitive accuracy of 76.11\% with a remarkably reduced training time of 16.2993 minutes, emphasizing {\papername}'s efficiency by being over 600 times faster than Glyph and presenting quicker computation than HETAL. This disparity underscores the effectiveness and adaptability of {\papername} across diverse data for image classification.

\subsection{ViT Vs DEiT}

The introduction of the Vision Transformer (ViT) model in 2020 marked a departure from conventional image classification architectures, relying solely on attention mechanisms while omitting convolutional neural networks \cite{dosovitskiy2020image}. ViT showcased commendable performance when pre-trained on expansive datasets, yet its training process proved computationally demanding. To address this challenge, the Data-Efficient Image Transformer (DEiT) emerged in 2021, specifically designed to enhance ViT's efficiency \cite{touvron2021training}. DEiT incorporates novel techniques like distillation and progressive resizing, strategically curbing ViT's data demands. Consequently, DEiT enables more resource-efficient model pre-training. Notably, DEiT models pre-trained on datasets, like ImageNet, achieve performance levels equivalent to ViT models when fine-tuned for tasks like object detection and segmentation \cite{touvron2021training}. Furthermore, fine-tuning DEiT models requires significantly fewer computational resources and training samples compared to their ViT counterparts. This efficiency renders DEiT an advantageous choice for fine-tuning various computer vision tasks, especially in scenarios constrained by limited training data. Given these advantages, we adopt DEiT in our implementation. Through our experiments, we observe that encrypted training using DEiT outperforms ViTs across all datasets, showcasing superior accuracy improvements and reduced training times.

\begin{table*}[!ht]
\caption{The comparison of {\papername} with respect to overall training latency}
\centering
\begin{tabular}{cccc}
\toprule
\textbf{Methodology} & \textbf{Dataset} & \textbf{Accuracy} & \textbf{Time} \\
\midrule
\multirow{4}{*}{Glyph \cite{lou2020glyph}} & MNIST & 98.6\% & 8 days\\
 & CIFAR-10 & Not reported & Not reported \\
 & Face Mask Detection & Not reported & Not reported\\
 & DermaMNIST & 73.2\% & 7 days \\
\midrule
 \multirow{4}{*}{HETAL \cite{lee2023hetal}} & MNIST & 96.73\%& 57.3715 minutes \\
 & CIFAR-10 & 96.57\% & 51.905 minutes \\
 & Face Mask Detection &  95.46\% & 9.4453 minutes\\
 & DermaMNIST & 76.06\% & 18.9498 1minutes\\
 \midrule
 \multirow{4}{*}{{\papername} (Ours)} & MNIST & 97.75\% & 37.839 minutes \\
 & CIFAR-10 & 95.20\% & 31.3342 minutes\\
 & Face Mask Detection & 97.18\% & 3.9086 minutes \\
 & DermaMNIST & 76.11 \%  & 16.2993  minutes\\
\bottomrule
\end{tabular}
\label{table:comparison}
\end{table*}

\section{Related Work}
Fully Homomorphic Encryption (FHE) \cite{gentry2009fully}, widely recognized as a holy grail of cryptography, has garnered considerable attention in privacy-preserving machine learning research. Graepel et al. introduced an innovative approach to binary classification confidentiality in their work, ML Confidential \cite{mlconfidential}, rooted in polynomial approximations and leveraging FHE. Simultaneously, Dowlin et al. explored inference and learning on encrypted data in CryptoNets \cite{gilad2016cryptonets}, based on FHE principles and modifications to activation functions, influencing advancements in privacy-preserving cloud-based neural networks.

Additionally, foundational work by Takabi et al. \cite{takabi2016privacy} in ensuring privacy in multiparty machine learning using FHE laid the groundwork for the evolution of privacy-preserving Deep Learning (DL) models. Hesamifard et al. \cite{hesamifard2018privacy} introduced a technique for approximating activation functions using low-degree polynomials, applied to Convolutional Neural Networks (CNNs). Alongside FHE and polynomial approximations, the Faster CryptoNets framework \cite{chou2018faster} demonstrated the efficacy of pruning and quantization in harnessing sparse representations within the cryptosystem, enhancing DL model inference performance.

In large-scale real-time data analysis, Han et al. \cite{han2019logistic} presented a logistic regression model grounded in FHE, showing promising results. Notably, Nandakumar et al. \cite{nandakumar2019towards} devised an FHE-based approach for training neural networks on homomorphically encrypted data, expanding the realm of privacy-preserving machine learning. The HomoPAI framework \cite{homopai}, proposed by Li et al., offers a secure collaborative Machine Learning (ML) service for processing data from diverse owners. Enriched with parallelism through the Message Passing Interface (MPI) framework, it achieves performance enhancements and time-saving attributes. Lou et al. introduced an approach for transfer learning using fast fully homomorphic encryption over the torus and BGV cryptosystems in \cite{lou2020glyph}. Recently, Lee et al. \cite{lee2023hetal} introduced privacy-preserving transfer learning using FHE, achieving faster performance while utilizing GPUs for computation.

\section{Conclusion}
In this work, we introduce {\papername}, a privacy-preserving fine-tuning system that leverages fully homomorphic encryption in conjunction with data-efficient image transformers. Our empirical findings underscore that {\papername} proficiently trains transformer models, maintaining parity in accuracy with unencrypted models. Furthermore, our experimental observations substantiate the superior performance of {\papername} when compared to prior approaches. In our future work, we will focus on extending our methodology to encompass more intricate computer vision datasets and large-scale transformer models.



\bibliography{aaai24}

\end{document}